\documentclass[10pt,twocolumn,letterpaper]{article}

\usepackage[final]{cvpr}

\usepackage{multicol}

\usepackage{graphicx}
\usepackage{amsmath}
\usepackage{amssymb}
\usepackage{booktabs}
\usepackage{multirow}
\usepackage{float}

\usepackage{pifont}
\newcommand{\cmark}{\ding{51}}%
\newcommand{\xmark}{\ding{55}}%

\usepackage[pagebackref,breaklinks,colorlinks]{hyperref}

\usepackage{color}
\usepackage{colortbl}
\usepackage{comment}
\usepackage{bbm}

\newcommand{\figref}[1]{Fig. \ref{#1}}
\newcommand{\tabref}[1]{Table \ref{#1}}

\newcommand*\samethanks[1][\value{footnote}]{\footnotemark[#1]}
\makeatletter
\def\hlinewd#1{%
\noalign{\ifnum0=`}\fi\hrule \@height #1 \futurelet
\reserved@a\@xhline}

\usepackage[capitalize]{cleveref}
\crefname{section}{Sec.}{Secs.}
\Crefname{section}{Section}{Sections}
\Crefname{table}{Table}{Tables}
\crefname{table}{Tab.}{Tabs.}

\def\cvprPaperID{2554}
\def\confName{CVPR}
\def\confYear{2022}

\begin{document}

\title{Semi-Supervised Learning of Semantic Correspondence with Pseudo-Labels}

\author{
    Jiwon Kim$^{1}$\thanks{Equal contribution} \quad Kwangrok Ryoo$^{1}$\samethanks \quad Junyoung Seo$^{1}$\samethanks \quad 	Gyuseong Lee$^{1}$\samethanks\\ \quad 	Daehwan Kim$^{2}$ \quad Hansang Cho$^{2}$ \quad Seungryong Kim$^{1}$\thanks{Corresponding author}\\
    $^{1}$Korea University, Seoul, Korea   $^{2}$Samsung Electro-Mechanics, Suwon, Korea\\
    {\tt\small \{naancoco,kwangrok21,se780,jpl358,seungryong\_kim\}@korea.ac.kr  } \\
    {\tt\small  \{daehwan85.kim,hansang.cho\}@samsung.com  } 
}
\maketitle

\begin{abstract}
Establishing dense correspondences across semantically similar images remains a challenging task due to the significant intra-class variations and background clutters. Traditionally, a supervised learning was used for training the models, which required tremendous manually-labeled data, while some methods suggested a self-supervised or weakly-supervised learning to mitigate the reliance on the labeled data, but with limited performance. 

In this paper, we present a simple, but effective solution for semantic correspondence that learns the networks in a semi-supervised manner by supplementing few ground-truth correspondences via utilization of a large amount of confident correspondences as pseudo-labels, called SemiMatch. Specifically, our framework generates the pseudo-labels using the model’s prediction itself between source and weakly-augmented target, and uses pseudo-labels to learn the model again between source and strongly-augmented target, which improves the robustness of the model. We also present a novel confidence measure for pseudo-labels and data augmentation tailored for semantic correspondence. In experiments, SemiMatch achieves state-of-the-art performance on various benchmarks.  
\end{abstract}

\begin{figure}[t]
  \begin{subfigure}[h]{0.233\textwidth}
    \includegraphics[width=\textwidth]{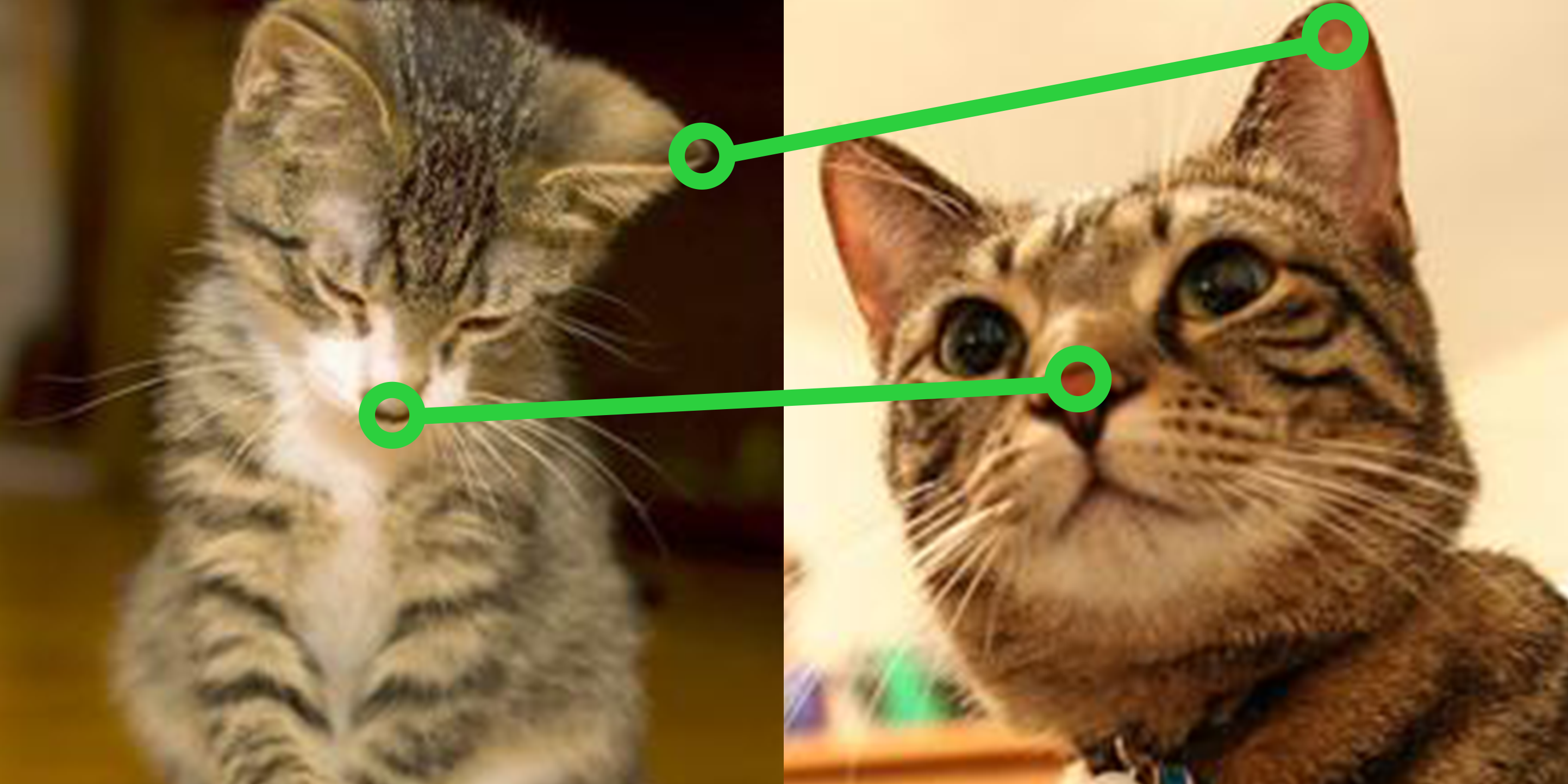}
    \caption{Supervised loss}
    \label{fig:sup}
  \end{subfigure}
  \begin{subfigure}[h]{0.233\textwidth}
    \includegraphics[width=\textwidth]{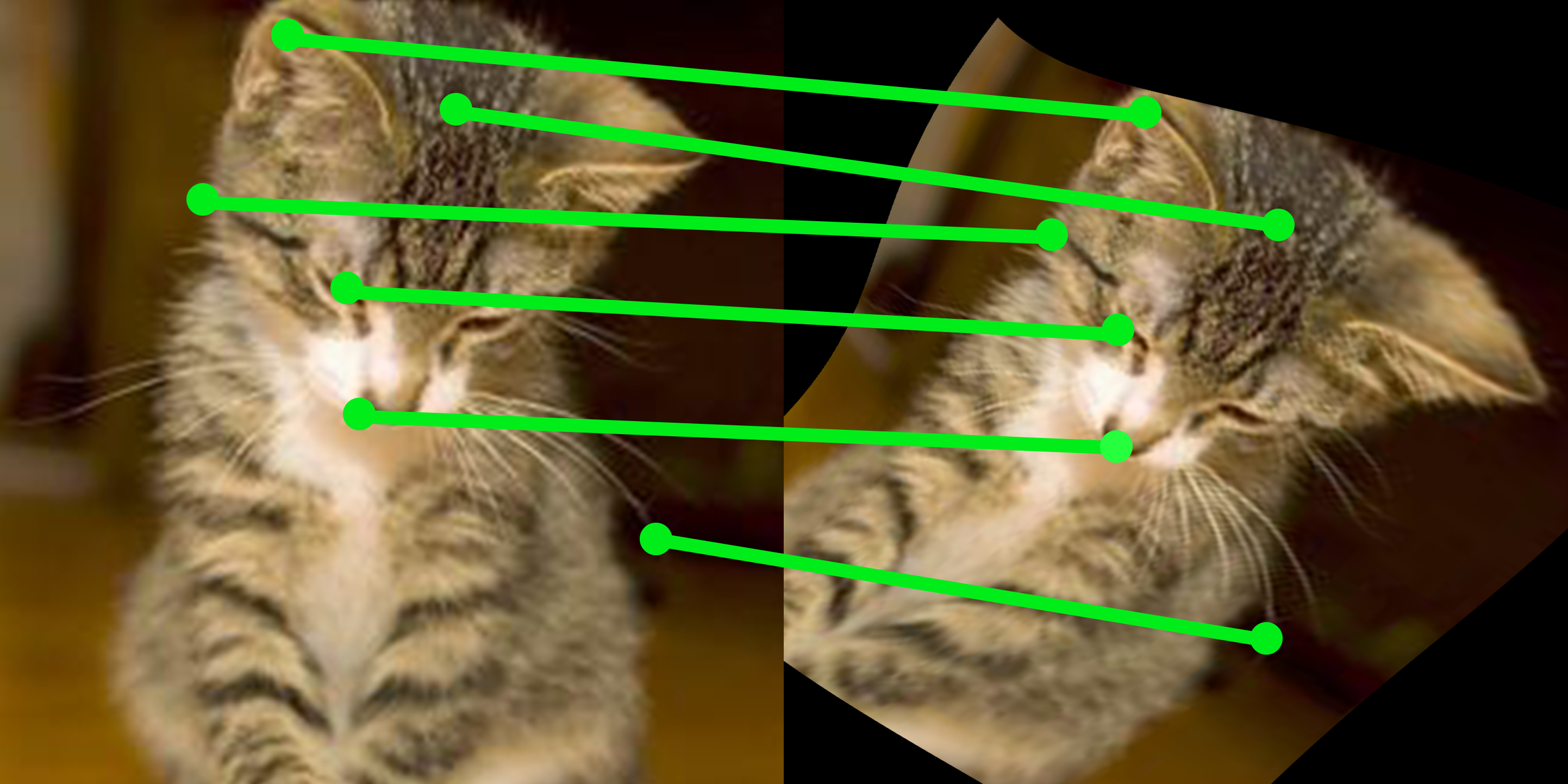}
    \caption{Self-supervised loss}
    \label{fig:self}
  \end{subfigure}
 \\
    \begin{subfigure}[h]{0.233\textwidth}
    \includegraphics[width=\textwidth]{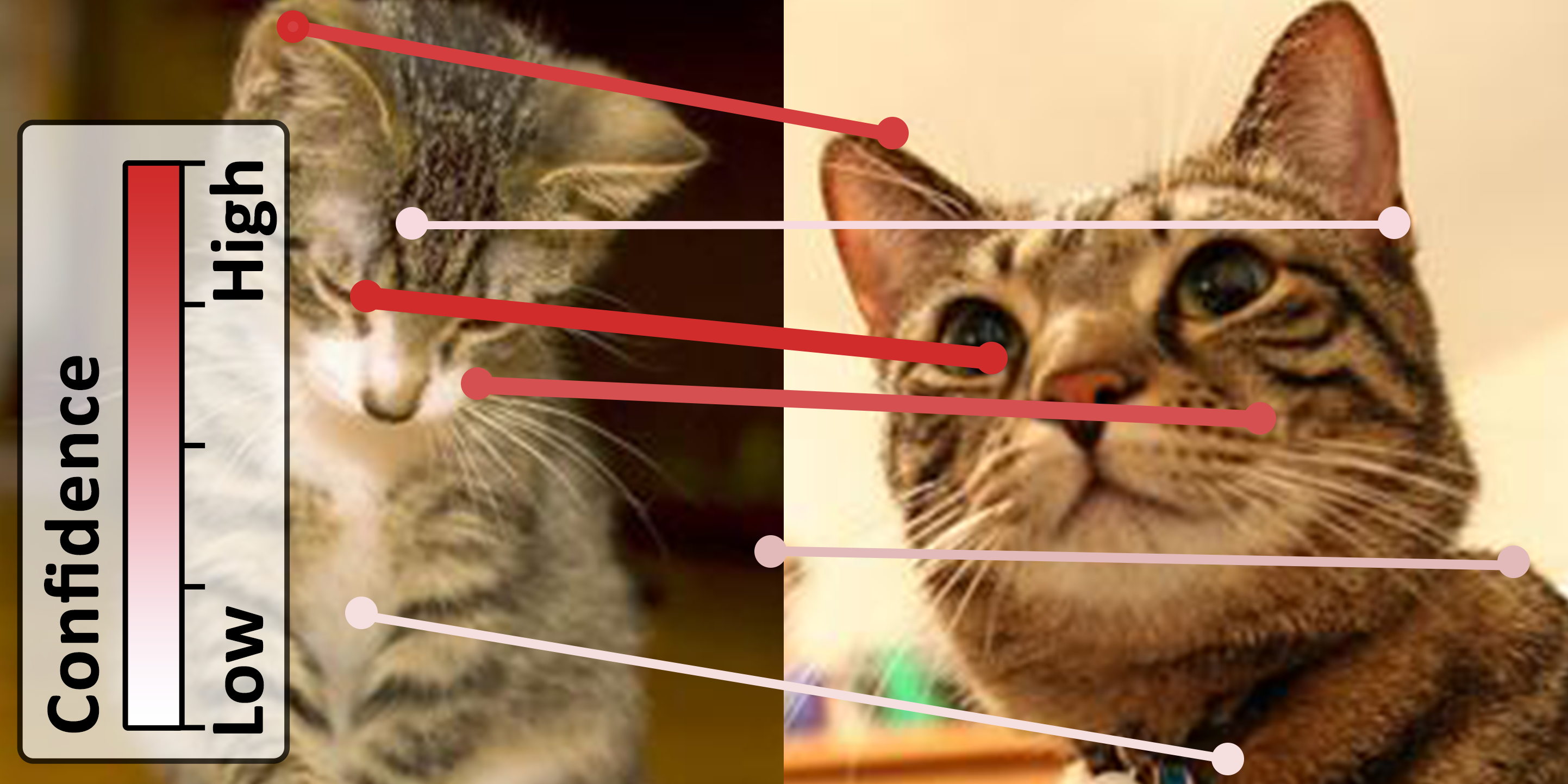}
    \caption{Unsupervised loss (Ours) }
    \label{fig:unsup}
  \end{subfigure}
     \begin{subfigure}[h]{0.233\textwidth}
    \includegraphics[width=\textwidth]{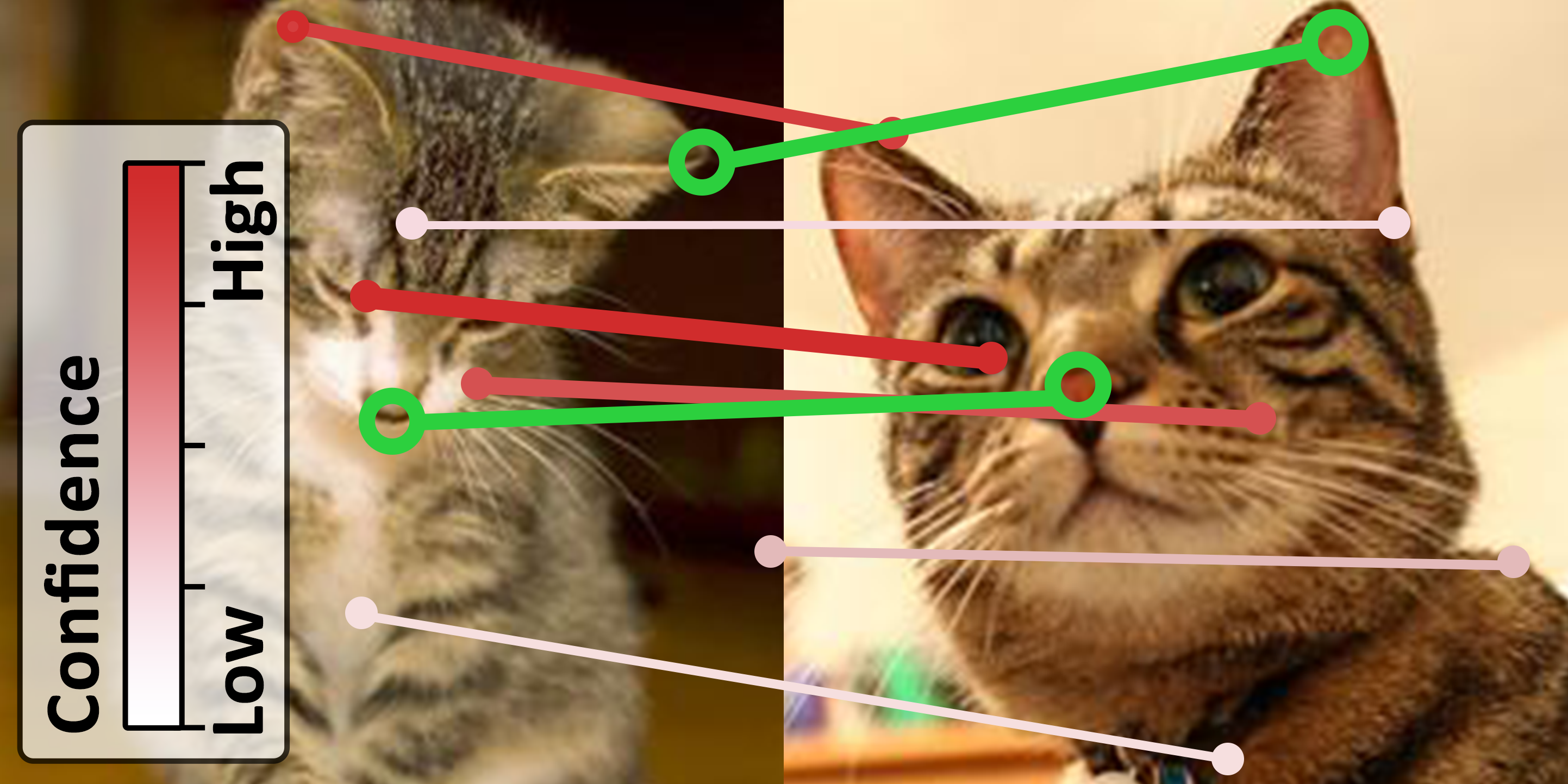}
    \caption{Semi-supervised loss (Ours) }
    \label{fig:semi}
  \end{subfigure}
  \vspace{-5pt}
  \caption{\textbf{Comparison of semantic correspondence methods in terms of different formulation of supervision.} Conventional methods leverage either (a) supervised loss using sparse ground-truth keypoint matches~\cite{han2017scnet,min2019hyperpixel,min2020learning} or (b) self-supervised loss using synthetic flow field with random geometric parameters~\cite{rocco2017convolutional, melekhov2019dgc, truong2020glu}. Unlike them, we present (c) unsupervised loss using pseudo-labels from a matching probability and (d) semi-supervised loss using both sparse ground-truth keypoints and confident pseudo-labels.}
  \label{fig:short}\vspace{-10pt}
\end{figure}

\section{Introduction}
Establishing dense correspondences across semantically similar images, depicting different instances of the same object or scene category, can facilitate many Computer Vision applications such as semantic segmentation~\cite{rubinstein2013unsupervised, taniai2016joint, min2021hypercorrelation}, object detection~\cite{lin2017feature}, or image editing~\cite{liao2017visual, kim2019semantic}. Unlike classical dense correspondence problems such stereo matching or optical flow~\cite{sun2018pwc,hui2018liteflownet}, semantic correspondence poses additional challenges from large intra-class appearance and geometric variations~\cite{choy2016universal, han2017scnet, kim2018recurrent}.

Although formulated in various ways, most recent approaches~\cite{rocco2017convolutional,rocco2018end,rocco2020ncnet, melekhov2019dgc,min2019hyperpixel,min2020learning,liu2020semantic,truong2020glu,sarlin2020superglue,truong2020gocor,min2021convolutional,cho2021semantic} addressed these challenges by carefully designing deep neural networks, such as CNNs~\cite{rocco2017convolutional,rocco2018end,rocco2020ncnet, melekhov2019dgc,min2019hyperpixel,min2020learning,liu2020semantic,truong2020glu,sarlin2020superglue,truong2020gocor} or Transformers~\cite{sun2021loftr,cho2021semantic}, based models. The most straightforward way to formulate a mapping function is to use ground-truth correspondences between the image pairs. Recent approaches~\cite{cho2021semantic,zhao2021multi}, including conventional approaches, have been formulated in a \textit{supervised} fashion (\figref{fig:sup}). However, ground-truth keypoint pairs on the most standard benchmarks~\cite{ham2017proposal,min2019spair} can be the inherent bottleneck~\cite{choy2016universal, han2017scnet,min2019hyperpixel,min2020learning} because they are annotated subjectively and sparsely.

To alleviate the reliance on the ground-truth data, some methods~\cite{rocco2017convolutional, melekhov2019dgc, truong2020glu, truong2020gocor} presented a \textit{self-supervised} learning framework (\figref{fig:self}), using synthetic geometric warps of an image to generate a synthetic image pair. Although it turns out that it is an appealing alternative, using synthetic image pairs cannot account for extreme intra-class appearance variations in semantic correspondence~\cite{rocco2017convolutional, seo2018attentive}. On the other hand, some other methods~\cite{kim2018recurrent,huang2019dynamic,kim2019semantic,min2020learning} presented a \textit{weakly-supervised} learning framework that casts this task as a feature reconstruction between the images, but the loss function often fails to explain the correspondences across severely different instances among the same class.

On the other hand, most recent approaches in image classification task~\cite{berthelot2019mixmatch, berthelot2019remixmatch, sohn2020fixmatch, kuo2020featmatch, hu2021simple} have been popularly formulated in a \textit{semi-supervised} learning framework, which enables learning the model on a large amount of \textit{unlabeled} data with a few \textit{labeled} data, and showed outstanding performance. Most recent trends of semi-supervised learning~\cite{berthelot2019mixmatch, berthelot2019remixmatch, sohn2020fixmatch, kuo2020featmatch, hu2021simple} integrate consistency regularization~\cite{bachman2014learning} and pseudo-labeling~\cite{lee2013pseudo}. For instance, FixMatch~\cite{sohn2020fixmatch} first generates a pseudo-label using the model's prediction on weakly-augmented unlabeled data and then encourages the prediction from strongly-augmented unlabeled data to follow the pseudo-label with confidence thresholding. This learning framework has become a promising solution to mitigate the reliance on large labeled data~\cite{hestness2017deep, mahajan2018exploring,sohn2020fixmatch}, but directly applying these techniques to semantic correspondence is challenging in that learning the matching networks requires pixel-level pseudo-labels.

\begin{figure*}[!t]
\begin{center}
\includegraphics[width=1\linewidth]{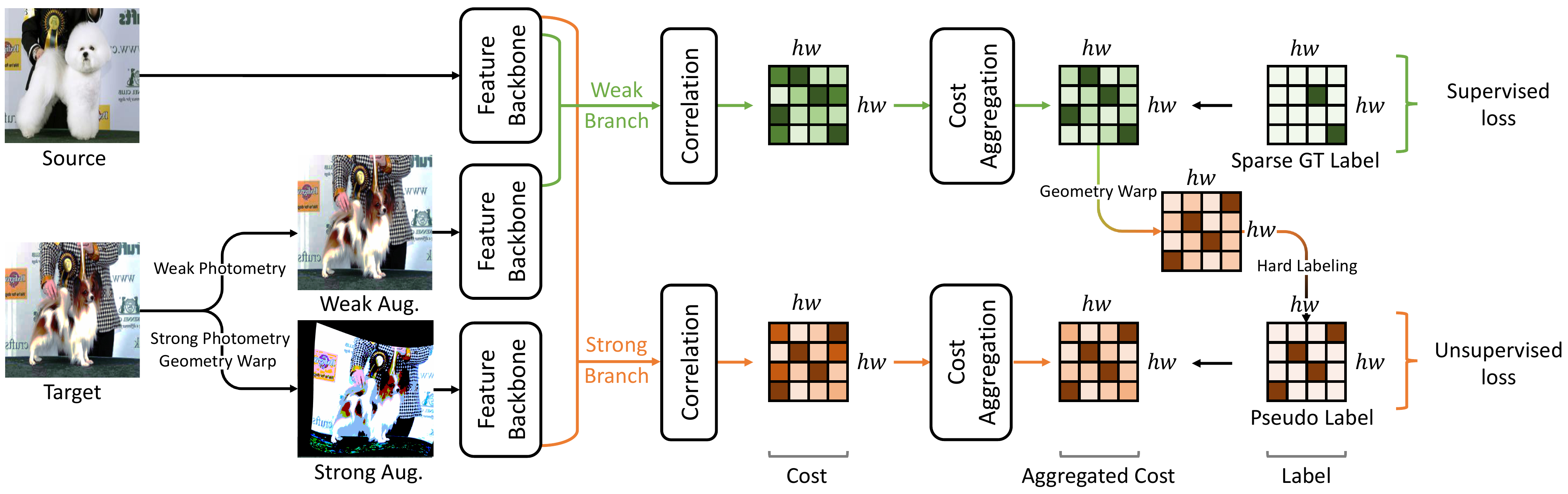}
\end{center}
\vspace{-10pt}
\caption{\textbf{Overview of our semi-supervised learning framework for semantic correspondence.} SemiMatch first augments the target image into a \textit{weakly-augmented} one and a \textit{strongly-augmented} one. The backbone matching networks extract features from these images, compute correlation cost, followed by aggregation. The aggregated cost between source and weakly-augmented target images is transformed by the same geometry warp used in strong augmentation and sharpened to generate a \textit{pseudo-label}. The overall loss function is composed of supervised loss which uses the sparse ground-truth label and unsupervised loss which uses the pseudo-label.}
\label{network_overall}\vspace{-10pt}
\end{figure*}

In this paper, we present a novel \textit{semi-supervised} learning framework, called SemiMatch, that generates pixel-level pseudo-labels using the model’s prediction itself between source and \textit{weakly-augmented} target and then encourages the model to predict the pseudo-label again between source and \textit{strongly-augmented} target, as illustrated in~\figref{network_overall}. To account for the observation that all of the pseudo-labels may not help to boost performance, we introduce a novel confidence measure for pseudo-labels by considering object-centric foreground, forward-backward consistency, and uncertainty of probability itself. Tailored for semantic correspondence, we also present a matching-specialized augmentation that exploits a keypoint-aware cutout, helping the network to learn distinctive feature representations around the keypoints.

We evaluate our method on several benchmarks~\cite{ham2017proposal, min2019spair}. Experimental results on various benchmarks show that using our novel loss function with sparsely-supervised loss function consistently improves the performance compared to the latest methods for semantic correspondence. We also provide an extensive ablation study to validate and analyze components in our learning framework.

\section{Related works}
\paragraph{Semantic Correspondence.} The objective of semantic correspondence~\cite{choy2016universal,ham2017proposal,kim2017fcss,rocco2018end} is to find correspondences across semantically similar images. 
Recent methods~\cite{choy2016universal, han2017scnet,min2019hyperpixel,min2020learning,cho2021semantic, zhao2021multi} showed great progress by supervised loss, but they are limited by the availability of sparse ground-truth annotation. On the other hand, some methods~\cite{rocco2017convolutional, melekhov2019dgc, truong2020glu, truong2020gocor} address the aforementioned limitations by training the network in a self-supervised manner, relying on synthetic warps of real images, without ground-truth flow and \cite{li2021probabilistic} presents a teacher-student model where the student model exploits the teacher model's generalized knowledge learned from synthetic data. Their improvement is mainly due to self-supervised training data, but synthetic geometric deformation cannot model intra-class appearance variations and realistic scene generation. Another alternative is to use a training loss that only requires a weak-level of supervision, given for each image pair as either positive (the same class) or negative (different class), as in~\cite{huang2019dynamic, rocco2020ncnet, min2020learning}. While it~\cite{min2020learning} can achieve higher performance than strongly-supervised methods, it has fundamental limitations to supervise locating matches. Unlike the methods above, we present for the first time a semi-supervised learning framework which overcomes the lack of labeled ground-truth keypoints by utilizing a large amount of confident correspondences as pseudo-labels.
\vspace{-10pt}

\paragraph{Semi-Supervised Learning.} 
The most recent methods for semi-supervised learning  have followed two trends; pseudo-labeling and consistency regularization. Pseudo-labeling~\cite{lee2013pseudo,shi2018transductive, arazo2020pseudo, xie2019unsupervised, zoph2020rethinking, pham2021meta} encourages a model to follow the pseudo-label from the model's prediction itself closely related to entropy minimization~\cite{grandvalet2005semi} where the model's predictions are encouraged to be low-entropy (i.e., high-confidence) on unlabeled data. 
On the other hand, consistency regularization~\cite{bachman2014learning, sajjadi2016regularization, tarvainen2017mean, xie2019unsupervised} encourages the model to produce the same prediction when perturbations are applied to the input or the model. Consistency regularization-based methods, enforcing invariant representations across augmentations, rely heavily on the usage of strong data augmentation, so it is important which strong augmentations are used and how strong they are. Very recently, some state-of-the-art methods~\cite{xie2019unsupervised, berthelot2019remixmatch, sohn2020fixmatch, kuo2020featmatch, hu2021simple} combine pseudo-labeling and consistency regularization by a confidence-based strategy and separate weak and strong augmentations, but there still remains the problem of ignoring a large amount of unlabeled data due to reliance on fixed high thresholding to compute the unsupervised loss. 
Most aforementioned methods have focused on solving an image-level task, e.g., image classification, but methods for pixel-level tasks such as semantic correspondence are done limitedly and cannot be applied directly. \vspace{-10pt}

\paragraph{Uncertainty Estimation.}
Predicting uncertainty in the field of Computer Vision has been widely explored~\cite{bruhn2006confidence, kybic2011bootstrap,mac2012learning,kondermann2007adaptive}, even before the expansion of deep learning. There are two approaches in uncertainty estimation: empirical and predictive. The former method ~\cite{graves2011practical, blundell2015weight} approximates the uncertainty by sampling a finite number of weight configurations for a given network and computing the mean and variance of the predictions. In the latter method~\cite{nix1994estimating}, a network is trained to infer the mean and variance of the distribution by design.

In semi-supervised learning in image classification, there were some approaches, such as Dash~\cite{xu2021dash} and FlexMatch~\cite{zhang2021flexmatch}, to measure the uncertainty of pseudo-labels by applying a different threshold to each sample, unlike FixMatch~\cite{sohn2020fixmatch} which assumes that the samples exceed the hand-crafted confidence threshold. In contrast to the existing semi-supervised method~\cite{sohn2020fixmatch,xu2021dash,zhang2021flexmatch} which assumes that all pseudo-labels are certain without considering the uncertainty of the generated pseudo-label, we improve the performance by considering the uncertainty of the pseudo-label.

\section{Methodology}
\subsection{Motivation}
Given a pair of images, i.e., source $I_s$ and target $I_t$, which represent semantically similar images, the goal of semantic correspondence is to establish matches between the two images at each pixel. To achieve this, most predominant methods with CNNs~\cite{rocco2017convolutional,rocco2018end,rocco2020ncnet, melekhov2019dgc,min2019hyperpixel,min2020learning,liu2020semantic,truong2020glu,sarlin2020superglue,truong2020gocor} consist of two steps, including feature extraction and cost aggregation. First of all, feature extraction networks extract a feature $F \in  \mathbb{R}^{h \times w \times d}$, where $h \times w$ is the spatial resolution and $d$ is the channels. The similarities between feature maps, called cost volume, can be estimated by 
$\mathcal{C}(i,j)=F_t(i)^T F_s(j)$, where $i \in \{1,..,h_t  w_t\}$ and $j \in \{1,..,h_s w_s\}$
However, initial cost volume itself is vulnerable to ambiguous, repetitive, or textureless matches. To disambiguate these, recent methods~\cite{liu2020semantic, li2020correspondence, min2021convolutional, cho2021semantic} employ the cost aggregation networks for refining the initial matching similarities to achieve the aggregated cost $\mathcal{C}'(i,j)$.

Although existing methods~\cite{liu2020semantic, li2020correspondence, min2021convolutional,cho2021semantic, zhao2021multi} can be formulated in various ways, their outputs $\mathcal{C}$ or $\mathcal{C}'$ can be considered as a \textit{matching probability} through a simple SoftMax function~\cite{lee2019sfnet} such that $P(i) = p(I_s, I_t(i);\theta) \in \mathbb{R}^{h_s w_s \times 1}$ with the network parameters $\theta$, defined across all the points in $I_s$ for point $i$ in $I_t$. Learning such networks in a \textit{supervised} manner requires manually annotated ground-truth correspondences $P_\mathrm{GT}$, which are extremely labor-intensive and involves subjectivity~\cite{min2020learning,cho2021semantic, zhao2021multi}. Thus, in semantic correspondence, only sparsely-annotated ground-truths are available, and the supervised loss function is defined such that
\begin{equation}\label{equ:loss_unsup}
    \mathcal{L}_\mathrm{sup} = \sum_{i}
    c(i){\mathcal{D}(p(I_s, I_t(i) ; \theta), P_\mathrm{GT}(i))},
\end{equation}
where $c(i)$ is a binary indicator for representing the existence of ground-truth $P_\mathrm{GT}(i)$, $P_\mathrm{GT}(i)$ is an one-hot vector form, and $\mathcal{D}(\cdot,\cdot)$ is the distance function, e.g., L2 distance~\cite{melekhov2019dgc} or cross-entropy~\cite{kim2018recurrent, rocco2020ncnet}. Due to the inherent nature of using sparse ground-truths, the distance function yields a limited performance~\cite{cho2021semantic, zhao2021multi}. 
To alleviate the reliance on large ground-truth data, \textit{self-supervised} learning methods~\cite{rocco2017convolutional, melekhov2019dgc, truong2020glu, truong2020gocor} have been popularly used, which generates synthetic matching pairs by applying a geometric warping on a single image. In specific, target image $I_t$ (or source image $I_s$) is transformed by geometry warping operator $\mathcal{G}(\cdot;\phi)$ with a randomly-defined warping field $\phi$, e.g., generated by affine ~\cite{weisstein2004affine} or thin-plate-spline (TPS)~\cite{bookstein1993thin} transformation. The synthetic image pairs are then defined as $I_t$ and $\mathcal{G}(I_t;\phi)$, and $\phi$ is used as a pseudo-label for them.

The self-supervised loss function is then defined as 
\label{equ:loss_self}
\begin{equation}
    \mathcal{L}_\mathrm{self-sup} = \sum_{i}
    \mathcal{D}(p(\mathcal{G}(I_t;\phi),I_t(i);\theta), P_\phi(i)).
\end{equation}
where $P_\phi$ is one-hot vector form of $\phi$. Since this loss function does not require annotated ground-truths, and enables learning with dense labels, it can be an alternative to data hungry from sparse annotations~\cite{cho2021semantic,zhao2021multi}. However, such synthetic image pairs cannot contain realistic appearance variations across the two images and model moving objects or occlusion, which limits the performance in semantic correspondence that often poses challenges by intra-class appearance and shape variations~\cite{min2021convolutional, lee2019sfnet}.

\begin{figure}[t!]
    \begin{subfigure}[t]{0.115\textwidth}
        \centering
		\includegraphics[width=1\textwidth]{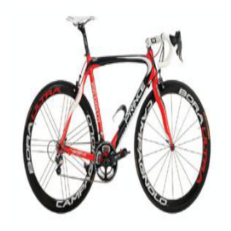}\hfill
        \caption{$I_s$}
    \end{subfigure}
    \centering
    \begin{subfigure}[t]{0.115\textwidth}
        \centering
		\includegraphics[width=1\textwidth]{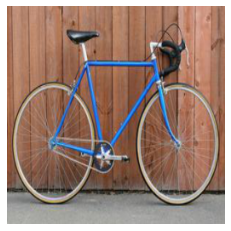}\hfill
        \caption{$I_t$}
    \end{subfigure}
    \begin{subfigure}[t]{0.115\textwidth}
        \centering
		\includegraphics[width=1\textwidth]{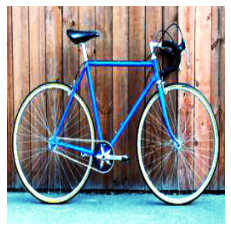}\hfill
        \caption{$\alpha(I_t)$}
    \end{subfigure}
    \begin{subfigure}[t]{0.115\textwidth}
        \centering
		\includegraphics[width=1\textwidth]{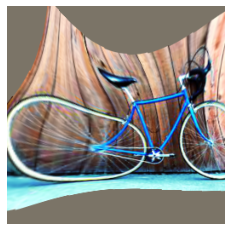}\hfill
        \caption{$\mathcal{G}(\mathcal{A}(I_t))$}
    \end{subfigure}
    \vspace{-10pt}
    \caption{\textbf{Examples of augmented images in semi-supervised learning:} (a) source image $I_s$, (b) target image $I_t$ (c) weakly-augmented target image $\alpha(I_t)$, and (d) strongly-augmented target image $\mathcal{G}(\mathcal{A}(I_t))$.}
\label{fig:geo_img}\vspace{-10pt}
\end{figure}

\subsection{Formulation}
We present a novel \textit{semi-supervised} learning framework, SemiMatch, for learning a matching model on a large amount of unlabeled pixels with few labeled pixels between source and target images. Following recent trends of semi-supervised learning in image classification~\cite{xie2019unsupervised, sohn2020fixmatch} that have not been applied in semantic correspondence yet, we present to extend the consistency regularization between two differently augmented instances from the same image~\cite{sohn2020fixmatch} to semantic correspondence. They are commonly based on the assumption that when perturbations are applied to the input, the prediction should not change significantly. However, it is difficult to adapt the existing consistency regularization techniques~\cite{berthelot2019mixmatch,berthelot2019remixmatch,sohn2020fixmatch} to semantic correspondence directly, predicting dense probabilities because they were designed for image classification task.

In this section, we study how to formulate the loss function for \textit{unsupervised learning} within semantic correspondence framework that can be simultaneously used with sparsely supervised loss function for semi-supervised learning. First of all, as shown in ~\figref{fig:geo_img} given source $I_s$ and target $I_t$ images, we build a triplet $\{I_s,\alpha(I_t),\mathcal{G}(\mathcal{A}(I_t))\}$, where $\alpha(\cdot)$ and $\mathcal{A}(\cdot)$ represent \textit{weak} and \textit{strong} photometric and geometric augmentations, respectively. Our key ingredient is to exploit the difference in difficulty levels of matching weak pairs and strong pairs, i.e., the weak pairs are easier to generate pseudo-labels for the strong pairs, inspired by ~\cite{sohn2020fixmatch}. However, without the geometric augmentation, the performance of trained models may be sub-optimal because the models may lack the robustness to geometric variations, which frequently occur in semantic correspondence. But if the geometric augmentation is applied simultaneously, direct consistency regularization cannot be formed,i.e., to
perform consistency regularization on the geometrically warped
image pairs, it is necessary to find the pseudo labels
in the misaligned image pair.

To overcome the aforementioned issues, we present a novel unsupervised loss function that jointly leverages photometric and geometric augmentations, defined such that 
\begin{equation}\label{equ:loss_sup}
    \mathcal{L}_\mathrm{un-sup} = \sum_{i}
    m(i) \mathcal{D}(p(I_s,\mathcal{G}(\mathcal{A}(I_t(i));\phi); \theta)
    , Q(i)),
\end{equation}
where $Q(i)$ is a pseudo-label defined as
\begin{equation}
    Q(i)=\mathcal{G}(p(I_s,\alpha(I_t(i));\theta);\phi),
\end{equation}
which means the geometrically-warped matching probabilities between weak pairs with $\phi$. And the same geometric warping applied to the target image, such that $\mathcal{G}(\mathcal{A}(I_t);\phi)$. Like previous consistency regularization methods~\cite{bachman2014learning}, we generate pseudo-labels from better correspondences of the weak pairs than the strong pairs as shown in ~\figref{fig:WeakStrong}. By geometrically-warping the pseudo-labels, we can align matching probability for strong pairs, enabling the model to achieve robustness to strong augmentation as well.

In the following section, we will explain how to achieve the confidence $m(i)$ of pseudo-label.

\subsection{Confidence of Pseudo-Label}
Matching probabilities inferred from the weak branch are used to define pseudo-labels in the strong branch. However, incorrect pseudo-labels may hinder performance boosting, which is called confirmation bias~\cite{arazo2020pseudo} problem. To overcome this, some semi-supervised learning methods for image classification use simple thresholding of the probability value itself, e.g., FixMatch~\cite{sohn2020fixmatch}. This is too a strong constraint in early iterations of training in that most false positives cannot be captured. 
 
To alleviate this, we present three constraints to measure the confidence of pseudo-labels.
\newcommand\mysize{0.23}
\begin{figure}[t]
  \centering
  \begin{subfigure}[b]{\mysize\textwidth}
    \centering
    \includegraphics[width=\linewidth]{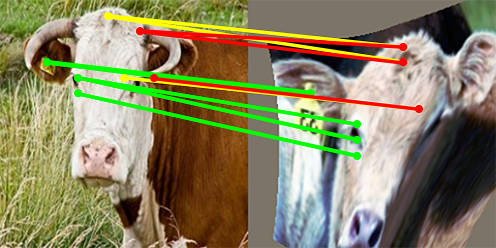}
  \end{subfigure}
    \begin{subfigure}[b]{\mysize\textwidth}
    \centering
    \includegraphics[width=\linewidth]{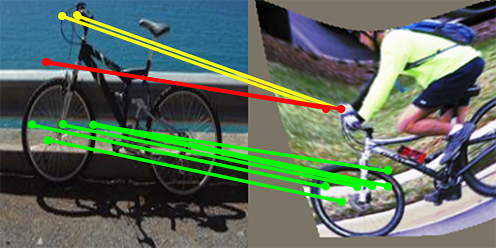}
  \end{subfigure}
   \caption{\textbf{Comparison of matching difficulty between weak branch and strong branch.}The yellow and red lines show the result of the weak and strong branch, respectively. The green lines show when the weak branch and strong branch results are the same. Weak branch gives more accurate results.
   }
   \label{fig:WeakStrong}\vspace{-10pt}
\end{figure}
First of all, we utilize a binary object mask of source and target images, which helps to limit the matching candidates within object-centric regions. In specific, since the pseudo-labels are defined at geometrically-warped regions, we define $M_\mathrm{mask}$ as a binary foreground object mask for target image $I_t$. Note that the object mask is synthesized using the min and max values of ground-truth keypoints.

Secondly, we also utilize forward-backward consistency checking~\cite{sundaram2010dense,meister2018unflow,liu2019ddflow}. In specific, by estimating correspondences between $\{I_s,\alpha(I_t)\}$ in forward and backward directions simultaneously and checking the consistency between them, we determine the reliability of pseudo-labels based on the observation that if this consistency constraint is not satisfied, the points in $\alpha(I_t)$ are occluded at the matches in $I_s$ (or vice-versa), and the estimated flow vector is incorrect. This constraint is denoted by $M_\mathrm{fb}$.

Finally, unlike existing consistency regularization methods~\cite{sohn2020fixmatch}, using a few samples through high thresholding with a pre-defined scalar value $\tau$, we present uncertainty-based weighting on the loss function itself. This lets a considerable amount of unlabeled data, especially at the early iteration of training phase, be used as pseudo-labels according to their different learning status. Specifically, we measure the uncertainty $u(i)$ of matching probabilities $P(i)$ as  
\begin{equation}
 u(i) = 1 / \exp\left(\sum\nolimits_{j}P(i,j)\log P(i,j)\right),
\end{equation}
where $P(i,j)$ is $j$-th component of $P(i)$.
$M_\mathrm{thres}(i)$ is then measured by $(1/u(i)) \odot 1\left(\max \left(P(i)\right) \geq \tau\right)$ to weight loss function depending on uncertainty where $\odot$ denotes Hadamard product.

\begin{figure}[t]
  \centering
  \begin{subfigure}[b]{0.15\textwidth}
    \centering
    \includegraphics[width=\textwidth]{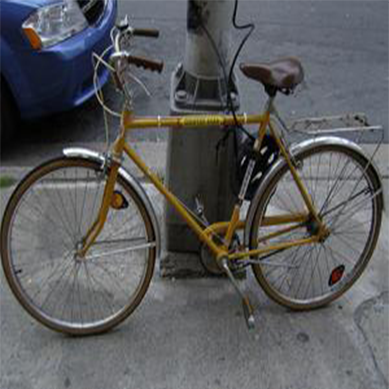}
       \caption{Source}
  \end{subfigure}
    \begin{subfigure}[b]{0.15\textwidth}
    \centering
    \includegraphics[width=\textwidth]{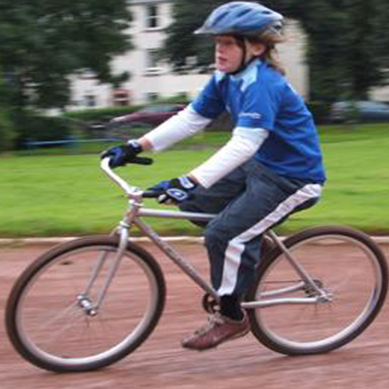}
         \caption{Target}  
  \end{subfigure} 
    \begin{subfigure}[b]{0.15\textwidth}
    \centering
    \includegraphics[width=\textwidth]{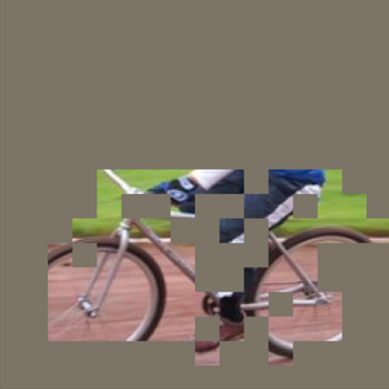}
           \caption{Mask intersection}
  \end{subfigure} 
  \\

  \begin{subfigure}[b]{0.15\textwidth}
    \centering
    \includegraphics[width=\textwidth]{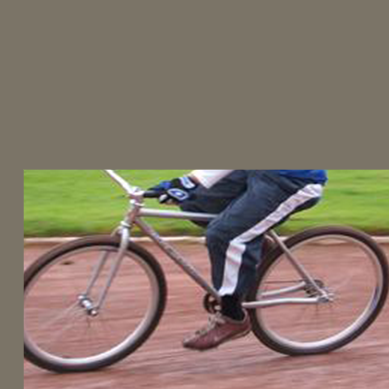}
  \caption{$M_\mathrm{mask}$}
  \end{subfigure}
    \begin{subfigure}[b]{0.15\textwidth}
    \centering
    \includegraphics[width=\textwidth]{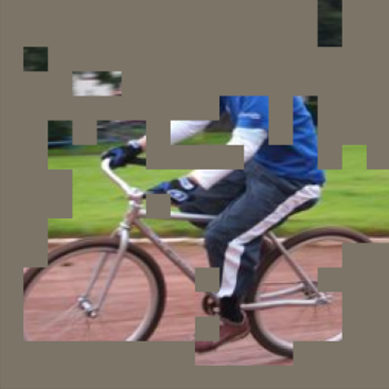}
   \caption{ $M_\mathrm{fb}$ }
  \end{subfigure} 
  \begin{subfigure}[b]{0.15\textwidth}
    \centering
    \includegraphics[width=\textwidth]{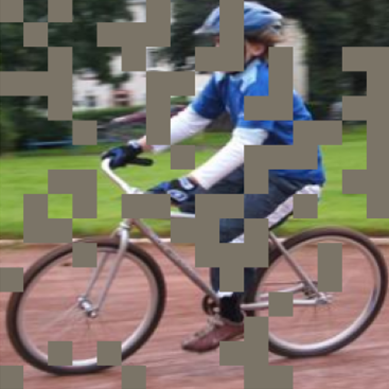}
   \caption{$M_\mathrm{thres}$ }
  \end{subfigure} \\
  \vspace{-5pt}
   \caption{\textbf{Visualization of our confidence mask:} (a) source, (b) target, (c) mask intersection, (d) $M_\mathrm{mask}$, (e) $M_\mathrm{fb}$, and (f) $M_\mathrm{thres}$. Sampled pixels by mask intersection belong to object-centric and discriminative image regions.}
   \label{fig:vis_mask}\vspace{-10pt}
\end{figure}

Our final confidence $m$ is determined by $\mathcal{G}(M_\mathrm{mask} \odot M_\mathrm{fb} \odot M_\mathrm{thres};\phi)$. 
We visualize each mask component belonging to the intersection of the masks in~\figref{fig:vis_mask}.

\subsection{Augmentation}
Consistency regularization-based methods~\cite{bachman2014learning}, enforcing invariant representations across augmentations, greatly depend on what kind of transformations are used for strongly-augmented images. Even though any kind of augmentation can be used in this framework, for semantic correspondence, augmentation should be more focused on discriminative local parts of an object to infer the matches.

\begin{figure}[t!]
        \begin{subfigure}[t]{0.16\textwidth}
        \centering
		\includegraphics[width=1\textwidth]{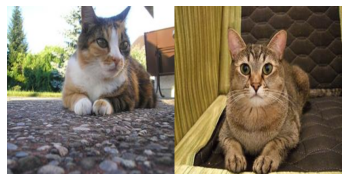}\hfill
        \caption{Image pair}
    \end{subfigure}%
    \centering
    \begin{subfigure}[t]{0.16\textwidth}
        \centering
		\includegraphics[width=1\textwidth]{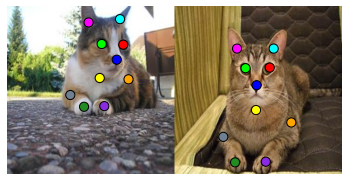}\hfill
        \caption{Keypoint pair}
    \end{subfigure}%
    \begin{subfigure}[t]{0.16\textwidth}
        \centering
		\includegraphics[width=1\textwidth]{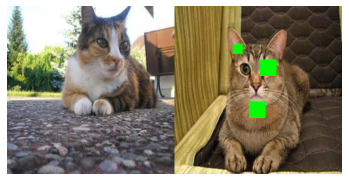}\hfill
        \caption{KeyOut pair}
    \end{subfigure}%
\vspace{-5pt}
    \caption{\textbf{Visualization of KeyOut augmentation:} (a) image pair, (b) keypoints, and (c) keyOut applied to the target image.}
\label{fig:keyout}\vspace{-10pt}
\end{figure}

In this section, we propose a matching-specialized augmentation, called keypoint-guided CutOut (KeyOut), which cuts and removes boxes of a certain size around the keypoint location as shown in ~\figref{fig:keyout}. It allows the model to learn to find the keypoint locations by integrating keypoint peripheral information.

\begin{table*}[t!]
\centering
\resizebox{0.9\textwidth}{!}{%
\begin{tabular}{l|c|c|ccc|ccc|c}
\hline
\multirow{2}{*}{Method} &
  \multirow{2}{*}{Supervision} &
  \multirow{2}{*}{\begin{tabular}[c]{@{}c@{}}Learning\\ signal\end{tabular}} &
  \multicolumn{3}{c|}{PF-PASCAL} &
  \multicolumn{3}{c|}{PF-Willow} &
  SPair-71k \\
          &                              &                                         & 0.05  & 0.1  & 0.15 & 0.05 & 0.1  & 0.15 & 0.1  \\ \hline
PF{\tiny{HOG}}~\cite{ham2017proposal}        & None                         & -                                       & 31.4  & 62.5 & 79.5 & 28.4 & 56.8 & 68.2 & -    \\ \hline
{CNNGeo{\tiny{ResNet-101}}~\cite{rocco2017convolutional}}    & \multirow{2}{*}{Self-sup.}   & \multirow{2}{*}{synthetic  pairs}       & 41.0  & 69.5 & 80.4 & 36.9 & 69.2 & 77.8 & 20.6 \\
{A2Net{\tiny{ResNet-101}}~\cite{seo2018attentive}}     &                              &                                         & 42.8  & 70.8 & 83.3 & 36.3 & 68.8 & 84.4 & 22.3 \\ \hline
{SF-Net{\tiny{ResNet-101}}~\cite{lee2019sfnet}}    & \multirow{6}{*}{Weak-sup.}   & bbox                                    & 53.6  & 81.9 & 90.6 & 46.3 & 74.0 & 84.2 & -    \\ \cline{1-1} \cline{3-10} 
{Weakalign{\tiny{ResNet-101}}~\cite{rocco2018end}} &                              & \multirow{5}{*}{images}           & 49.0  & 74.8 & 84.0 & 37.0 & 70.2 & 79.9 & 20.9 \\
{RTNs{\tiny{ResNet-101}}~\cite{kim2018recurrent}}      &                              &                                         & 55.2  & 75.9 & 85.2 & 41.3 & 71.9 & 86.2 & 25.7 \\
{NC-Net{\tiny{ResNet-101}}~\cite{rocco2020ncnet}}    &                              &                                         & 54.3  & 78.9 & 86.0 & 33.8 & 67.0 & 83.7 & 20.1 \\
{DCC-Net{\tiny{ResNet-101}}~\cite{huang2019dynamic}}   &                              &                                         & 55.6  & 82.3 & 90.5 & 43.6 & 73.8 & 86.5 & -    \\
DHPF{\tiny{ResNet-101}}~\cite{min2020learning}      &                              &                                         & 56.1  & 82.1 & 91.1 & 50.2 & \underline{80.2} & \underline{91.1} & 37.3 \\ \hline
SCNet{\tiny{VGG-16}}~\cite{han2017scnet}     & \multirow{7}{*}{Sup.} & \multirow{7}{*}{keypoints} & 36.2  & 72.2 & 82.0 & 38.6 & 70.4 & 85.3 & -    \\
ANC-Net{\tiny{ResNet-101-FCN}}~\cite{li2020correspondence}   &                              &                                         & -     & 86.1 & -    & -    & -    & -    & 28.7    \\
HPF{\tiny{ResNet-101}}~\cite{min2019hyperpixel}       &                              &                                         & 60.1  & 84.8 & 92.7 & 45.9 & 74.4 & 85.6 & 28.2 \\
DHPF{\tiny{ResNet-101}}~\cite{min2020learning}      &                              &                                         & 75.7  & 90.7 & 95.0 & 49.5 & 77.6 & 89.1 & 37.3 \\
CHMNet{\tiny{ResNet-101}}~\cite{min2021convolutional}  &                              &                                         & \textbf{80.1}  & 91.6 & 94.9 & \underline{52.7} & 79.4 & 87.5 & 46.3 \\

MMNet{\tiny{ResNet-101}}~\cite{zhao2021multi}     &                              &                                         & \underline{77.6}  & 89.1 & 94.3 & -    & -    & -    & 40.9 \\
CATs$^\dagger${\tiny{ResNet-101}}~\cite{cho2021semantic}      &                              &                                         & 67.5  & 89.1 & 94.9 & 46.6 & 75.6 & 87.5 & 42.4 \\
CATs{\tiny{ResNet-101}}~\cite{cho2021semantic} &                              &                                         & 75.4  & \underline{92.6} & \underline{96.4} & 50.3 & 79.2 & 90.3 & \underline{49.9} \\ \hline

\textbf{SemiMatch}$^\dagger$  &
  \multirow{2}{*}{Semi-sup.} &
  \multirow{2}{*}{\begin{tabular}[c]{@{}c@{}}keypoints \end{tabular}}     

                                           & 75.0     & 91.7 & 95.6    & 47.4    & 76.3    & 88.2    &  43.0    \\

\textbf{SemiMatch} &                              &                                         & \textbf{80.1} & \textbf{93.5} & \textbf{96.6} &\textbf{54.0}  & \textbf{82.1}    & \textbf{92.1}    & \textbf{50.7} \\ \hline
\end{tabular}%
}
\vspace{-5pt}
\caption{\textbf{Quantitative evaluation on PF-PASCAL and PF-Willow~\cite{ham2017proposal} and  SPair-71k~\cite{min2019spair}.} Subscripts of each method's name indicate the feature backbone used. The best results in bold, and the second best results are underlined. CATs$^\dagger$ means without fine-tuning feature.}
\label{tab:all}
\end{table*}

\subsection{Loss Functions}
Finally, we propose two loss functions to train our model using different supervision, including \textit{supervised} loss and \textit{semi-supervised} regimes. As described above, any distance function $\mathcal{D}$ can be used for loss functions. In specific, following the common practice~\cite{melekhov2019dgc}, $L_\mathrm{sup}$ is defined as the L2 distance such that
\begin{equation}
    \mathcal{L}_\mathrm{sup} = \sum_{i} c(i)\| \xi(
     p(I_s, I_t(i) ;\theta)) - \xi({P}_\mathrm{GT}(i)) \|,
\label{equ:overall_loss_sup}
\end{equation}
where $\xi(\cdot)$ is denoted as general max function including soft argmax and hard argmax. 
In addition, $L_\mathrm{un-sup}$ is formulated with the contrastive loss function~\cite{wang2021dense} as 
\begin{equation}
\begin{split}
&\mathcal{L}_\mathrm{un-sup} \\
&= -\sum_{i} m(i) \log \left(\frac{\exp \left( p(I_s(i'),G(i); \theta) / \gamma\right)}{\sum_{j} \exp (p(I_s(j),G(i); \theta) / \gamma)}\right),
\label{equ:lossUnsup} 
\end{split}
\end{equation}
where $G=\mathcal{G}(\mathcal{A}(I_t);\phi)$, $i'$ and $j$ represent the locations in source image, respectively, $i'$ is determined as $\xi({Q(i))}$, and $\gamma$ is the temperature hyper-parameter. 

Our total loss $\mathcal{L}_\mathrm{total} = \mathcal{L}_\mathrm{sup}+\lambda\mathcal{L}_\mathrm{un-sup}$ where $\lambda$ is a weight that is adaptively determined by the ratio between $\mathcal{L}_\mathrm{sup}$ and $\mathcal{L}_\mathrm{un-sup}$ such that $\lambda=\mathcal{L}^*_\mathrm{sup}/\mathcal{L}^*_\mathrm{un-sup}$, where $\mathcal{L}^*$ is the loss value itself and no back propagation happens.

\subsection{Network Architecture}
Our semi-supervised learning framework can be used in any deep networks for semantic correspondence~\cite{rocco2020ncnet, min2019hyperpixel, rocco2020efficient, jeon2020guided, liu2020semantic, li2020correspondence, min2021convolutional, cho2021semantic}. In this paper, we leverage the recent state-of-the-art network, especially focusing on cost aggregation stage, CATs~\cite{cho2021semantic} that explores global consensus among initial correlation map with the help of Transformer-based aggregator. By considering the outputs of the networks as matching probabilities, it directly leverages the proposed unsupervised loss, as well as supervised loss. Our overall architecture is shown in ~\figref{network_overall}.

\begin{table*}[]
\centering
\resizebox{\textwidth}{!}{%
\begin{tabular}{l|llllllllllllllllll|l}
\hlinewd{0.8pt}
Methods    & aero. & bike & bird & boat & bott. & bus  & car  & cat  & chai. & cow  & dog  & hors. & mbik. & pers. & plan. & shee. & trai. & tv   & all  \\ \hline
CNNGeo~\cite{rocco2017convolutional}     & 23.4  & 16.7 & 40.2 & 14.3 & 36.4  & 27.7 & 26.0 & 32.7 & 12.7  & 27.4 & 22.8 & 13.7  & 20.9  & 21.0  & 17.5  & 10.2  & 30.8  & 34.1 & 20.6 \\
A2Net~\cite{seo2018attentive}      & 22.6  & 18.5 & 42.0 & 16.4 & 37.9  & 30.8 & 26.5 & 35.6 & 13.3  & 29.6 & 24.3 & 16.0  & 21.6  & 22.8  & 20.5  & 13.5  & 31.4  & 36.5 & 22.3 \\
WeakAlign~\cite{rocco2018end}  & 22.2  & 17.6 & 41.9 & 15.1 & 38.1  & 27.4 & 27.2 & 31.8 & 12.8  & 26.8 & 22.6 & 14.2  & 20.0  & 22.2  & 17.9  & 10.4  & 32.2  & 35.1 & 20.9 \\
NC-Net~\cite{rocco2020ncnet}     & 17.9  & 12.2 & 32.1 & 11.7 & 29.0  & 19.9 & 16.1 & 39.2 & 9.9   & 23.9 & 18.8 & 15.7  & 17.4  & 15.9  & 14.8  & 9.6   & 24.2  & 31.1 & 20.1 \\
HPF~\cite{min2019hyperpixel}        & 25.2  & 18.9 & 52.1 & 15.7 & 38.0  & 22.8 & 19.1 & 52.9 & 17.9  & 33.0 & 32.8 & 20.6  & 24.4  & 27.9  & 21.1  & 15.9  & 31.5  & 35.6 & 28.2 \\
SCOT~\cite{liu2020semantic}       & 34.9  & 20.7 & 63.8 & 21.1 & 43.5  & 27.3 & 21.3 & 63.1 & 20.0  & 42.9 & 42.5 & 31.1  & 29.8  & 35.0  & 27.7  & 24.4  & 48.4  & 40.8 & 35.6 \\
DHPF~\cite{min2020learning}       & 38.4  & 23.8 & 68.3 & 18.9 & 42.6  & 27.9 & 20.1 & 61.6 & 22.0  & 46.9 & 46.1 & 33.5  & 27.6  & 40.1  & 27.6  & 28.1  & 49.5  & 46.5 & 37.3 \\
CHMNet~\cite{min2021convolutional} & 49.1  & 33.6 & 64.5 & \underline{32.7} & 44.6  & 47.5 & 43.5 & 57.8 & 21.0  & 61.3 & 54.6 & 43.8  & 35.1  & \textbf{43.7}  & 38.1  & \underline{33.5}  & 70.6  & 55.9 & 46.3 \\
MMNet~\cite{zhao2021multi}      & 43.5  & 27.0 & 62.4 & 27.3 & 40.1  & 50.1 & 37.5 & 60.0 & 21.0  & 56.3 & 50.3 & 41.3  & 30.9  & 19.2  & 30.1  & 33.2  & 64.2  & 43.6 & 40.9 \\
CATs$^\dagger$~\cite{cho2021semantic}       & 46.5  & 26.9 & 69.1 & 24.3 & 44.3  & 38.5 & 30.2 & 62.7 & 15.9  & 53.7 & 52.2 & 46.7  & 32.7  & 35.2  & 32.2  & 31.2  & 68.0  & 49.1 & 42.4 \\
CATs~\cite{cho2021semantic}     & \underline{52.0}  & \underline{34.7} & \underline{72.2} & \textbf{34.3} & \textbf{49.9}  & \underline{57.5} & \textbf{43.6} & \underline{66.5} & \textbf{24.4}  & \underline{63.2} & \underline{56.5} & \textbf{52.0}  & \underline{42.6}  & \underline{41.7}  & \underline{43.0}  & \textbf{33.6}  & \underline{72.6}  & \underline{58.0} & \underline{49.9} \\ \hline
\textbf{SemiMatch}$^\dagger$ & 47.8      & 29.0     & 70.6     & 24.0     & 44.5      & 37.6     & 29.8     & 65.2     & 17.2      & 54.7     & 52.8     & 47.1      & 35.2      & 37.6      & 29.9      & 32.7      & 68.5      & 49.4     & 43.0     \\ 
\textbf{SemiMatch} & \textbf{53.6}      & \textbf{37.0}     & \textbf{74.6}     & 32.3     & \underline{47.5}       & \textbf{57.7}     & 42.4      & \textbf{67.4}      & \underline{23.7}       & \textbf{64.2}      & \textbf{57.3}      & \underline{51.7}       & \textbf{43.8}      & 40.4       & \textbf{45.3}       & 33.1       & \textbf{74.1}       & \textbf{65.9}      & \textbf{50.7}

\\ \hlinewd{0.8pt}
\end{tabular}%
}
\vspace{-5pt}
\caption{\textbf{Per-class quantitative evaluation on SPair-71k dataset~\cite{min2019spair}.} The best results are in bold, and the second best results are underlined}
\label{tab:spair_detail}
\end{table*}

\section{Experiments}
\subsection{Implementation Details}

In experiments, we evaluate our framework with the state-of-the-art network, CATs~\cite{cho2021semantic}. For a fair comparison, we use the same hyper-parameters and photometric augmentation lists following CATs~\cite{cho2021semantic}. For geometric transformation~\cite{rocco2017convolutional, lee2019sfnet, truong2020glu}, we apply a combination of affine and thin-plate-spline with random transformation parameters in range [0,1] * $t_{scale}$ (0.15 for affine, and 0.4 for tps). To generate matching-specialized augmentations, we add blur as in \cite{chen2020improved} and KeyOut on strong augmentation. We set $\gamma=0.1$ and $\tau=0.5$.

\begin{figure*}[t!]
  \begin{subfigure}[b]{0.245\textwidth}
    \centering
    \includegraphics[width=\textwidth]{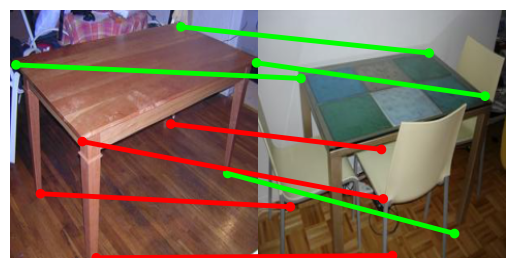}
  \end{subfigure}\hfill
  \begin{subfigure}[b]{0.245\textwidth}
    \centering
    \includegraphics[width=\textwidth]{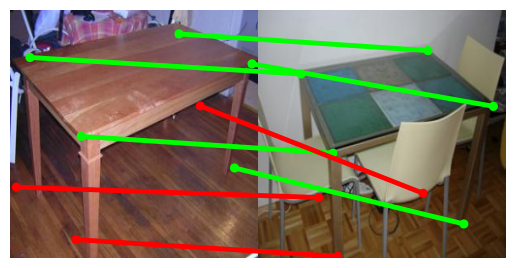}
  \end{subfigure}\hfill
  \begin{subfigure}[b]{0.245\textwidth}
    \centering
    \includegraphics[width=\textwidth]{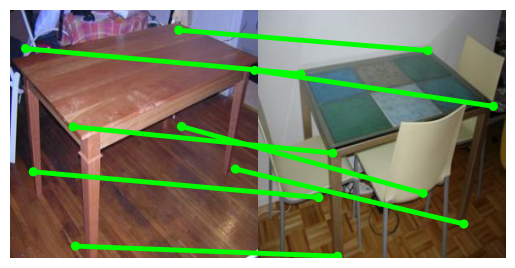}
  \end{subfigure}\hfill
    \begin{subfigure}[b]{0.245\textwidth}
    \centering
    \includegraphics[width=\textwidth]{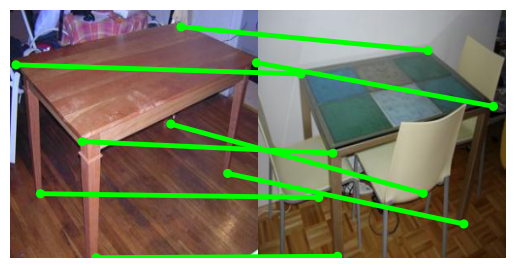}
  \end{subfigure}\hfill\\
  \begin{subfigure}[b]{0.245\textwidth}
    \centering
    \includegraphics[width=\textwidth]{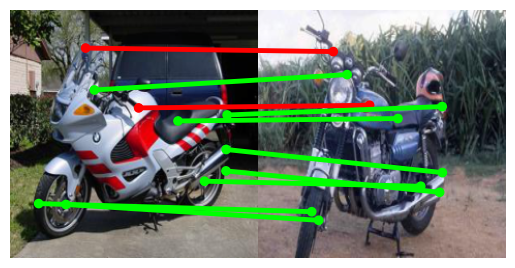}
    \caption{CHMNet~\cite{zhao2021multi}}
  \end{subfigure}\hfill
  \begin{subfigure}[b]{0.245\textwidth}
    \centering
    \includegraphics[width=\textwidth]{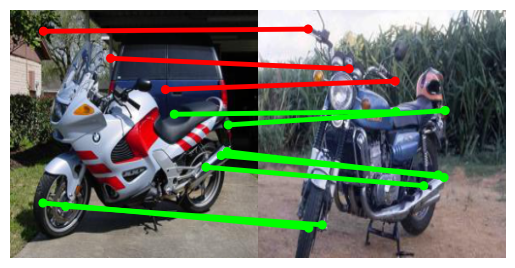}
   \caption{CATs~\cite{cho2021semantic}}
  \end{subfigure}\hfill
  \begin{subfigure}[b]{0.245\textwidth}
    \centering
    \includegraphics[width=\textwidth]{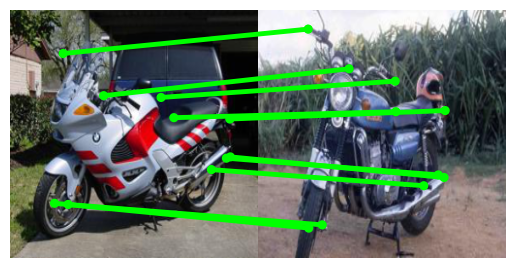}
  \caption{\textbf{SemiMatch}}
  \end{subfigure}\hfill
    \begin{subfigure}[b]{0.245\textwidth}
    \centering
    \includegraphics[width=\textwidth]{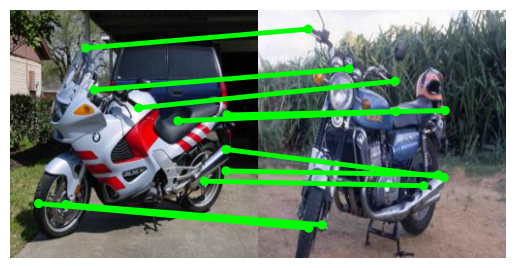}
   \caption{GT keypoints}
  \end{subfigure}\hfill\\
  \vspace{-15pt}
  \caption{\textbf{Qualitative results on PF-PASCAL~\cite{ham2017proposal}.} (a) CHMNet~\cite{min2021convolutional} (b) CATs~\cite{cho2021semantic} (c) SemiMatch, and (d) sparse GT keypoints.}
  \label{fig:qual}\vspace{-10pt}
\end{figure*} 

\subsection{Experimental Settings}

In this section, we conduct comprehensive experiments for semantic correspondence, by evaluating our framework through comparisons to state-of-the-art methods including HOG~\cite{ham2017proposal}, CNNGeo~\cite{rocco2017convolutional}, A2Net~\cite{seo2018attentive}, SFNet~\cite{lee2019sfnet}, WeakAlign~\cite{rocco2018end}, SCNet~\cite{han2017scnet}, RTNs~\cite{kim2018recurrent}, NC-Net~\cite{rocco2020ncnet}, DCC-Net~\cite{huang2019dynamic}, HPF~\cite{min2019hyperpixel}, DHPF~\cite{min2020learning}, SCOT~\cite{liu2020semantic}, ANC-Net~\cite{li2020correspondence}, CHM~\cite{min2021convolutional}, CATs~\cite{cho2021semantic}, MMNet~\cite{zhao2021multi}. 
\vspace{-10pt}
\paragraph{Dataset.}
We extensively conduct experiments on three popular benchmarks for semantic correspondence: PF-PASCAL~\cite{ham2017proposal}, PF-WILLOW~\cite{ham2017proposal}, SPair-71k~\cite{min2019spair}. PF-PASCAL contains 1,351 semantically related image pairs from 20 categories of the PASCAL VOC dataset~\cite{everingham2015pascal}. PF-Willow contains 900 image pairs from 4 categories. SPair-71k consists of total 70,958 image pairs in 18 categories with diverse view-point and scale variations. We used the same split proposed in~\cite{min2019spair}.
\vspace{-10pt}

\begin{table}[]
\centering
\resizebox{0.45\textwidth}{!}{%
\begin{tabular}{cl|cc|cc}
\hlinewd{0.8pt}
\multicolumn{2}{c|}{\multirow{2}{*}{Component}} & \multicolumn{2}{c|}{PF-PASCAL} & \multicolumn{2}{c}{PF-Willow} \\ \cline{3-6} 
\multicolumn{2}{c|}{}                         & 0.05 & 0.1 & 0.05 & 0.1 \\ \hline
(I)   & SemiMatch      & 80.1  & 93.5 & 54.0  & 82.1 \\
(II)  & (I) w/o $M_\mathrm{mask}$  &\underline{73.9}  & \underline{92.0} & \textbf{51.4}  & \textbf{79.7} \\
(III) & (I) w/o $M_\mathrm{fb}$   & 78.7  & 93.1 & 52.9  & 81.9 \\
(IV)  & (I) w/o $M_\mathrm{thres}$  & \textbf{72.3}  & \textbf{91.3} & \underline{52.1}  & \underline{81.5} \\ \hline
\hlinewd{0.8pt}
\end{tabular}%
}
\vspace{-5pt}
\caption{\textbf{Ablation study of mask elements.}}
\label{tab:ab_mask}\vspace{-10pt}
\end{table}

\paragraph{Evaluation Metric.}
Following the standard experimental protocol~\cite{ham2017proposal,min2019spair}, we use the percentage of correct keypoint (PCK@$\alpha$), computed as the ratio of estimated keypoints within the threshold from ground-truths to the total number of keypoints. Given a set of predicted and ground-truth keypoint pairs $\mathcal{K}=\left\{\left(k_{\text {pred}}(m), k_{\mathrm{GT}}(m)\right)\right\}$, PCK can be defined as $\operatorname{PCK}(\mathcal{K})=\frac{1}{M} \sum_{m}^{M} d\left(k_{\text {pred }}(m), k_{\mathrm{GT}}(m)\right) \leq \alpha_k \cdot \max (H, W)$, where $M$ is the number of keypoint pairs, $d(\cdot)$ is Euclidean distance; a threshold is scaled by $\alpha_k \cdot \max(H,W)$ in proportion to the larger portion of image for PF-PASCAL~\cite{ham2017proposal}, and the object's bounding box for PF-Willow~\cite{ham2017proposal}, and SPair-71k~\cite{min2019spair}.

\subsection{Matching Results}
\label{matching_results}
For fair comparisons with our baseline, CATs~\cite{cho2021semantic}, and previous state-of-the-art methods, we exploit the same network architecture, ResNet-101~\cite{he2016deep} pretrained on ImageNet~\cite{deng2009imagenet}. 
As shown in ~\tabref{tab:all}, for PF-PASCAL~\cite{ham2017proposal}, SemiMatch records state-of-the-art results with {80.1}\% PCK@{0.05}, {93.5}\% PCK@{0.1} and {96.6}\% PCK@0.15.  Without and with fine-tuning feature extraction backbone, SemiMatch outperforms CATs~\cite{cho2021semantic} by {7.5}\%/{4.7}\% PCK@0.05, {2.6}\%/{0.9}\% PCK@0.1, and {0.7}\%/{0.2}\% PCK@0.15. It demonstrates the effectiveness of our semi-supervised framework with the confidence constraints of pseudo-labels and matching-specialized augmentation. Compared to CATs, it can operate more sensitively at local regions through the significant performance improvement in PCK@0.05, which is the most strict matching criterion in PF-PASCAL. Experiments show that a large amount of pseudo-labels provide information on neighboring keypoints that cannot be provided by sparse keypoints. 
Generalization power of SemiMatch can be proven through the best performance in PF-willow for all PCKs by 1.3\%, 1.9\% and 1.0\%, respectively compared to the previous state-of-the-art results. Finally, we also record the best performance with 50.7\% PCK@0.1 even in SPair-71k~\cite{min2019spair} having large-scale variation. To show the effectiveness and robustness of our framework in detail, we compare per-class accuracy in~\tabref{tab:spair_detail} and our approach outperforms all state-of-the-art networks on 11 of the 18 classes. Our qualitative results are shown in~\figref{fig:qual}.

\subsection{Ablation Study} 
We conduct ablation analyses to investigate the effectiveness of components in our framework and also explore the effect of pseudo-labeling compared to CATs~\cite{cho2021semantic}. All experiments are conducted on PF-PASCAL dataset~\cite{ham2017proposal} and validated on PF-PASCAL and PF-Willow. \vspace{-10pt}
\begin{figure*}[t!]
        \begin{subfigure}[t]{0.244\textwidth}
        \centering
		\includegraphics[width=1\textwidth]{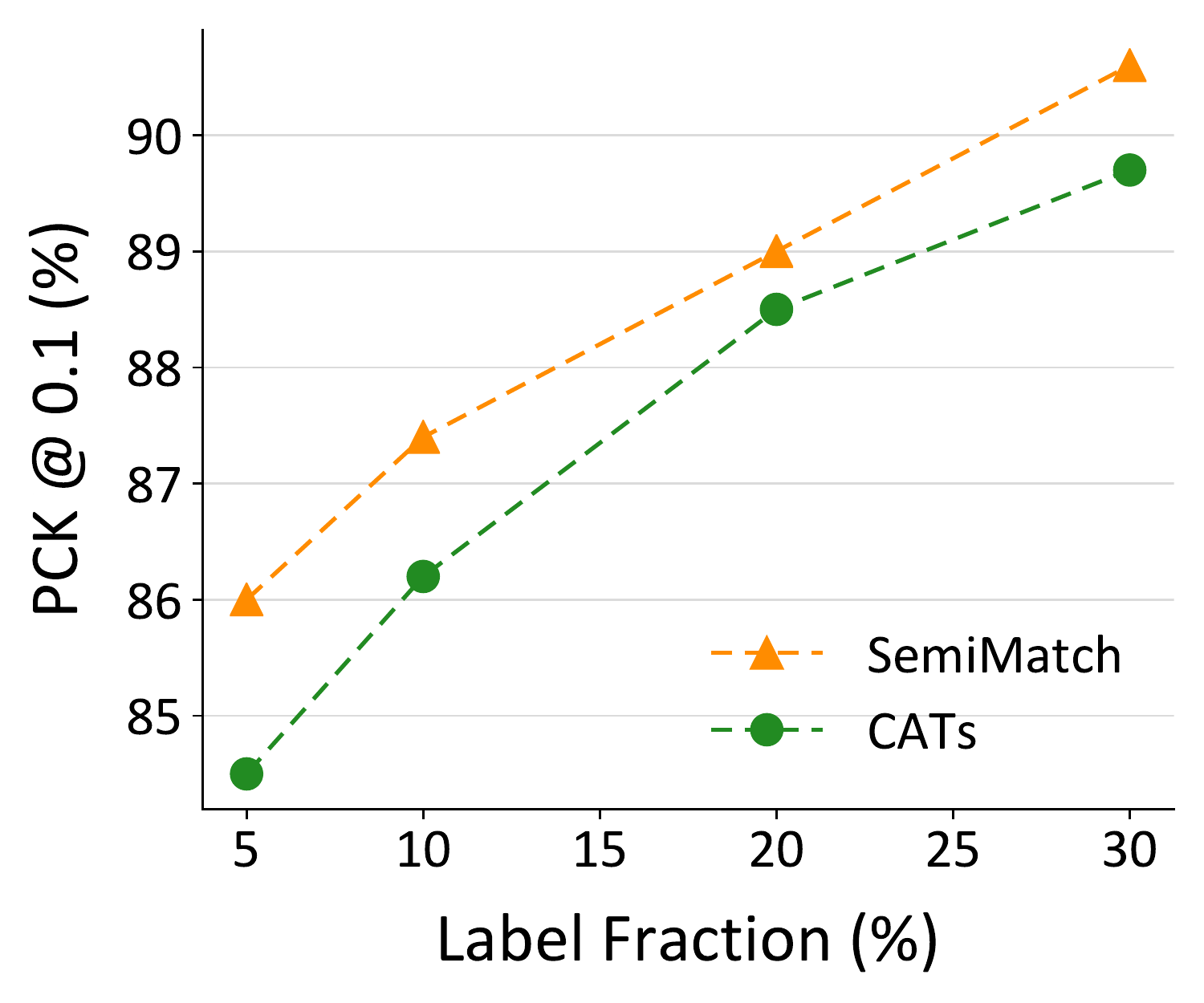}\hfill
        \caption{PCK@0.1 on PF-PASCAL}
    \end{subfigure}%
    \centering
    \begin{subfigure}[t]{0.244\textwidth}
        \centering
		\includegraphics[width=1\textwidth]{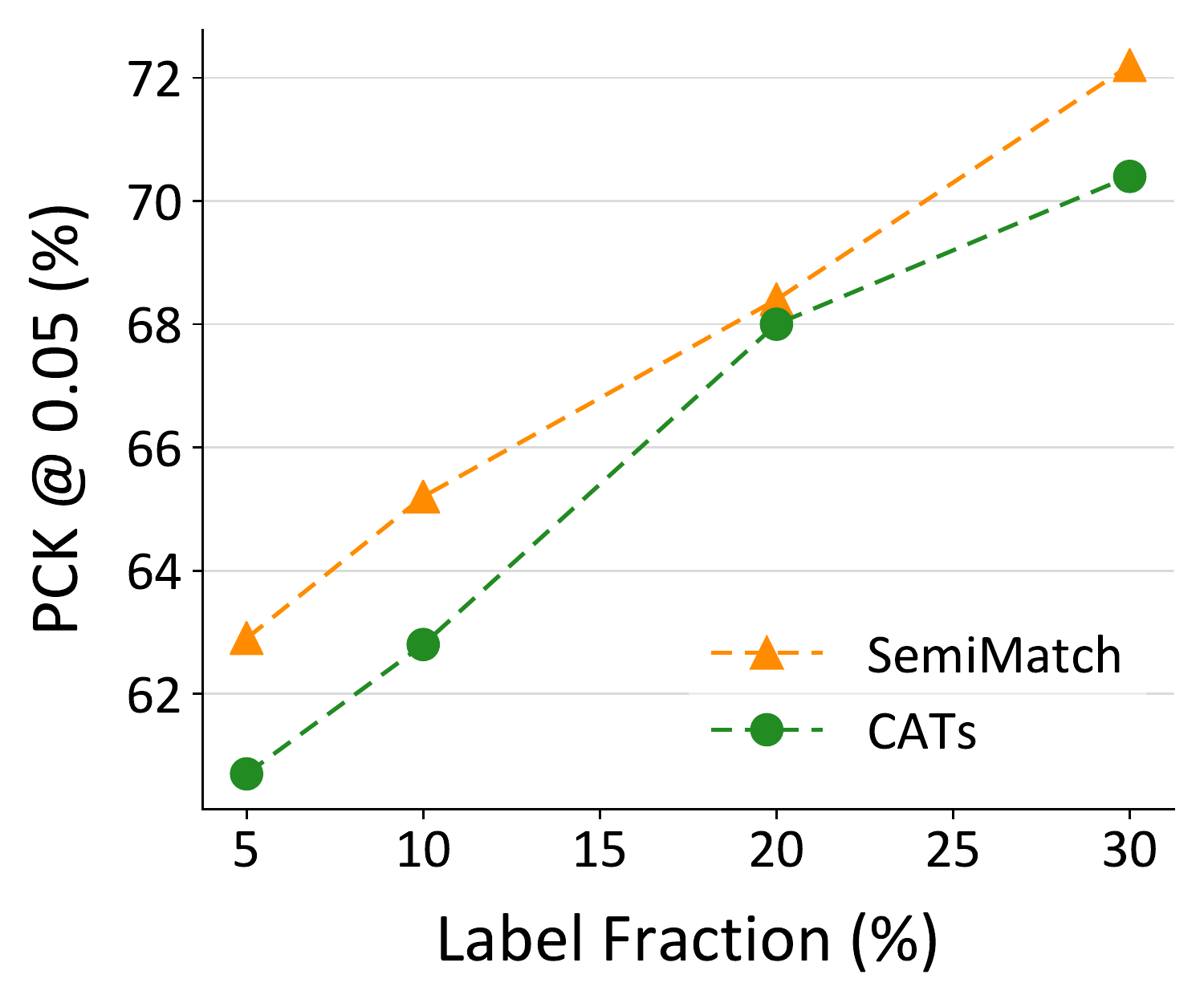}\hfill
        \caption{PCK@0.05 on PF-PASCAL}
    \end{subfigure}%
    ~ 
    \begin{subfigure}[t]{0.244\textwidth}
        \centering
		\includegraphics[width=1\textwidth]{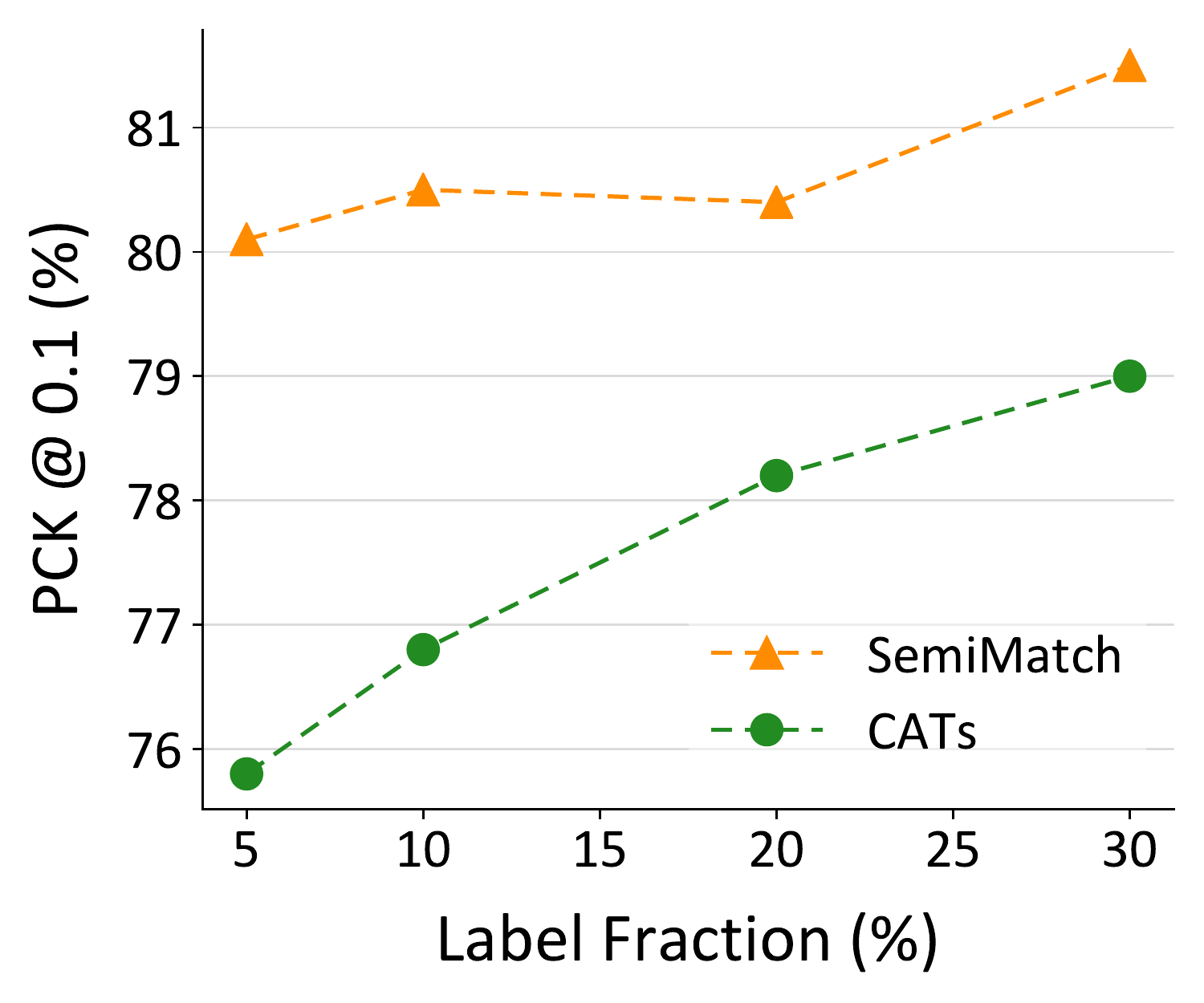}\hfill
        \caption{PCK@0.1 on PF-Willow}
    \end{subfigure}
        \begin{subfigure}[t]{0.244\textwidth}
        \centering
		\includegraphics[width=1\textwidth]{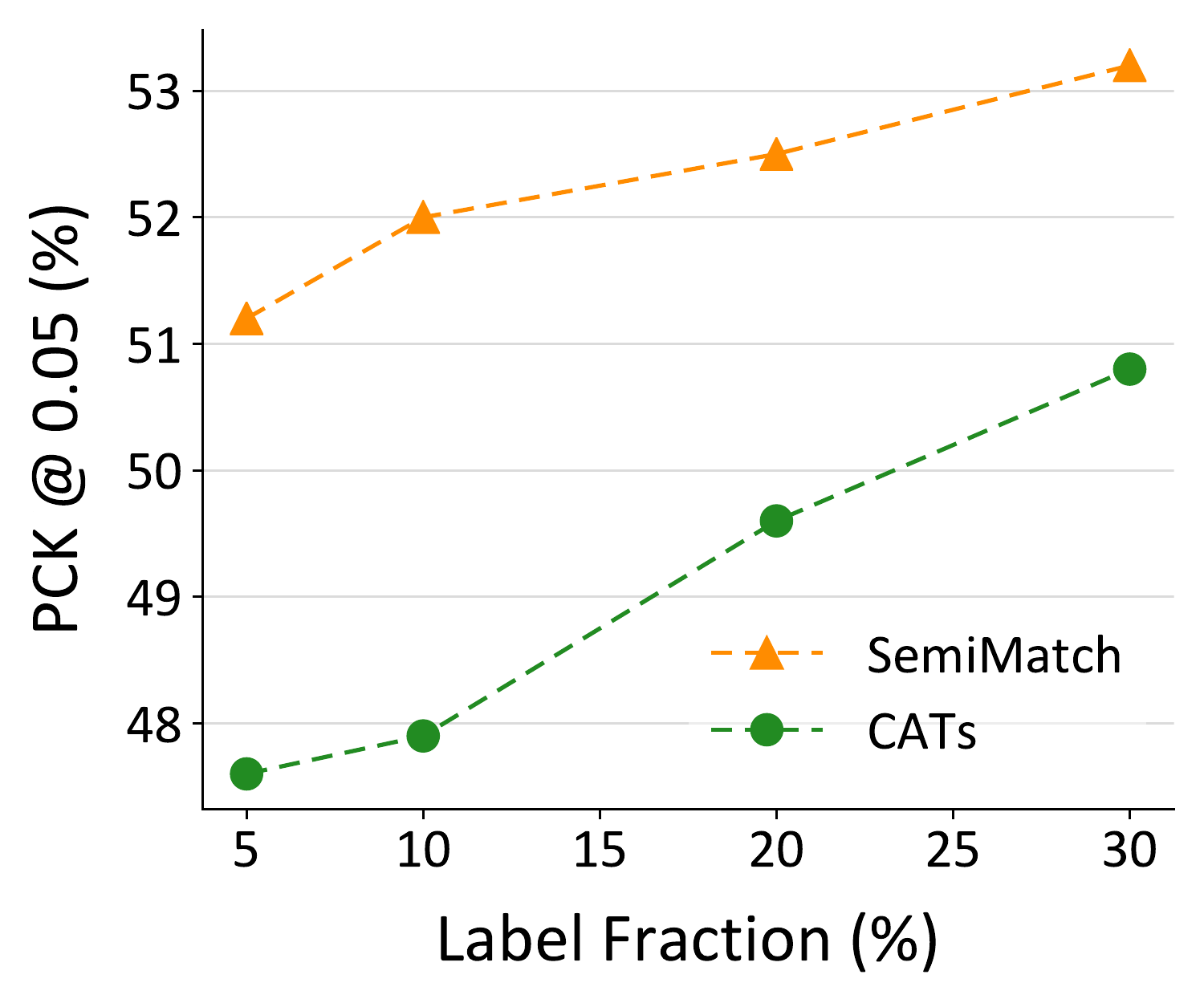}\hfill
        \caption{PCK@0.05 on PF-Willow}
        ~
    \end{subfigure}%

    \vspace{-10pt}
    \caption{{\textbf{PCK results of CATs~\cite{cho2021semantic} and SemiMatch on PF-PASCAL and PF-Willow with various label fractions:} (a) PCK@0.1 on PF-PASCAL, (b) PCK@0.05 on PF-PASCAL, (c) PCK@0.1 on PF-Willow, and (d) PCK@0.05 on PF-Willow.}}
    \label{fig:ab_label}
\end{figure*}

\paragraph{Effects of Confidence Mask.}
In~\tabref{tab:ab_mask}, we evaluate each mask constraint in SemiMatch baseline (I) by removing each from the entire constraints. From (I) to (II), PCK decline shows that the network can learn more representative information of the object by foreground samples separated from the background. A large performance drop in (IV) compared to (III) shows that it is important to adjust a weighting on the loss function proportionally according to the uncertainty. \vspace{-10pt}

\paragraph{Effects of Keypoint-Based Augmentation.}
We also evaluate additional strong augmentation by CutOut, KeyOut with low probability and small box regions ($\textrm{KeyOut}_{\tiny{\textrm{weak}}}$), and KeyOut with high probability and large box regions ($\textrm{KeyOut}_{\tiny{\textrm{strong}}}$). As shown in ~\tabref{tab:ab_cut}, we can prove the effectiveness of our KeyOut, cutout regions based on keypoint locations by comparing (II) and (III)-(IV). Especially on PF-Willow, (III) shows powerful generalization power compared to (I) and (II) by  1.1\% and 2.2\%.
\vspace{-10pt}

\begin{table}[]
\centering
\resizebox{0.45\textwidth}{!}{
\begin{tabular}{cl|cc|cc}
\hlinewd{0.8pt}
\multicolumn{2}{c|}{\multirow{2}{*}{Component}} & \multicolumn{2}{c|}{PF-PASCAL} & \multicolumn{2}{c}{PF-Willow} \\ \cline{3-6} 
\multicolumn{2}{c|}{}                         & 0.05 & 0.1 & 0.05 & 0.1 \\ \hline
(I)   & $\textrm{SemiMatch}_{\textrm{Base}}$      & \underline{79.8}  & 93.3 & 52.9  & 81.5 \\
(II)  & w/ $\textrm{CutOut}_{\textrm{weak}}$  &\underline{79.8}  & 93.2 & 51.8  & 81.1 \\
(III) & w/ $\textrm{KeyOut}_{\textrm{weak}}$  & \textbf{80.1}  & \textbf{93.6} & \textbf{54.0}  & \textbf{82.1} \\
(IV)  & w/ $\textrm{KeyOut}_{\textrm{strong}}$ & \underline{79.8}  & \underline{93.4} & \underline{53.6}  & \underline{81.9} \\ \hline

\hlinewd{0.8pt}
\end{tabular}
}
\vspace{-5pt}
\caption{\textbf{Ablation study of types of data augmentation.}}\vspace{-10pt}
\label{tab:ab_cut}
\end{table}

\paragraph{Experiments with Label Fraction.}
We investigate the performance gap between CATs and SemiMatch according to the label fraction of PF-PASCAL dataset, referring to the percentage of data in entire image pairs. We conduct ablation experiments using labels of 5\%, 10\%, 20\%, and 30\% of the total dataset.
~\figref{fig:ab_label} shows that SemiMatch is consistently $\mathtt{\sim}$4.5\% better than CATs in any label fraction setting on both PF-PASCAL and PF-Willow. Note that SemiMatch using 5\%, 10\%, 20\%, and 30\% of labels surpasses CATs using all labels by 0.9\%, 1.3\%, 1.2\% and 2.3\% PCK@0.1 for PF-Willow, respectively. Also, SemiMatch using 5\%, 10\%, 20\%, and 30\% of labels surpasses CATs using all labels by 0.9\%, 1.7\%, 2.2\%, and 2.9\% PCK@0.05 for PF-Willow.

\vspace{-10pt}

\paragraph{Experiments with Warm-Up stage.}
The fundamental problem of pseudo-labeling is that in the early stages of the training, the model is hindered by incorrect pseudo-labels caused by parameter initialization, which is called confirmation bias~\cite{arazo2020pseudo}. Therefore, semi-supervised frameworks based on pseudo-labeling generally have a warm-up stage using only the labeled data. We evaluate the effect of warm-up stage results, which is 20 epochs of training with supervised loss. In experiments, there is no significant difference in performance with or without the warm-up, as shown in~\tabref{tab:ab_warm}. It can be interpreted that our network explores the effectiveness of utilizing unlabeled data according to the model's learning status by uncertainty-based confidence measurement. \vspace{-10pt}
\begin{table}[]
\centering
\resizebox{0.9\linewidth}{!}{%
\begin{tabular}{c|ccc|ccc}
\hlinewd{0.8pt}
\multirow{2}{*}{Warm-up} & \multicolumn{3}{c|}{PF-PASCAL} & \multicolumn{3}{c}{PF-Willow} \\ \cline{2-7} 
                         & 0.05      & 0.1     & 0.15     & 0.05     & 0.1     & 0.15     \\ \hline
\cmark  & \textbf{80.1}          & \textbf{93.6}        & 96.6         & \textbf{54.0}         & 82.1        & 92.1         \\ \hline
\xmark  & 78.9          & 93.5        & \textbf{96.8}         & 53.9         & \textbf{82.3}        & \textbf{92.7}         \\ 
\hlinewd{0.8pt}
\end{tabular}%
}
\vspace{-5pt}
\caption{\textbf{Ablation study of warm-up stage.}}
\label{tab:ab_warm}
\end{table}

\begin{table}[]
\centering
\resizebox{0.9\linewidth}{!}{%
\begin{tabular}{c|ccc|ccc}
\hlinewd{0.8pt}
\multirow{2}{*}{Loss function} & \multicolumn{3}{c|}{PF-PASCAL} & \multicolumn{3}{c}{PF-Willow} \\ \cline{2-7} 
                         & 0.05 & 0.1 & 0.15 & 0.05 & 0.1 & 0.15 \\ \hline
AEPE  & 78.9 & 92.7 & 96.4 & 53.6 & \textbf{82.1} & \textbf{92.2}  \\ \hline
Contrastive  & \textbf{80.1} & \textbf{93.6} & \textbf{96.6}  & \textbf{54.0} & \textbf{82.1} &  92.1 \\ 
\hlinewd{0.8pt}
\end{tabular}%
}
\vspace{-5pt}
\caption{\textbf{Ablation study of loss function.}}
\label{tab:loss_comparison}\vspace{-10pt}
\end{table}
\paragraph{Comparison of Distance Function.}
A distance function for unsupervised loss in our semi-supervised framework can be defined as either AEPE or Contrastive loss. It should be noted that, SemiMatch shows better performance than CATs, our baseline, regardless of loss formulation. We experimentally observed that using Contrastive loss performs better than using AEPE as shown in ~\tabref{tab:loss_comparison}.

\section{Conclusion}
In this paper, we have presented a novel semi-supervised learning framework, called SemiMatch, that exploits the pixel-level pseudo-label generated by source and weakly-augmented target to learn a model again by taking source and strongly-augmented target as input. We introduce a novel confidence measure for pseudo-labels to ignore incorrect pseudo-labels and augmentation tailored for semantic matching, exploiting keypoint locations, to learn the model to integrate keypoint peripheral information. We have shown that SemiMatch achieves state-of-the-art performance over the latest methods in several benchmarks.

\noindent\textbf{Acknowledgements.}
This research was supported by the MSIT, Korea (IITP-2022-2020-0-01819, ICT Creative Consilience program), National Research Foundation of Korea (NRF-2021R1C1C1006897), and the Research and Development Program for Advanced Integrated Intelligence for Identification (AIID) (NRF-2018M3E3A1057288). 

{\small
\bibliographystyle{ieee_fullname}
\bibliography{egbib}
}

\end{document}